\documentclass[a4paper, 10pt, conference]{ieeeconf}      

\IEEEoverridecommandlockouts                              

\usepackage{graphics} 
\usepackage{color, colortbl}
\usepackage[ruled,linesnumbered]{algorithm2e}
\usepackage{epsfig} 
\usepackage{subfigure}
\usepackage{amsmath} 
\usepackage{amssymb}  
\usepackage{bm}
\usepackage{booktabs}
\usepackage{tabularx}
\usepackage{multirow}
\usepackage{graphicx}
\usepackage{xcolor}
\usepackage{soul}
\usepackage{leftidx}
\usepackage{comment}
\usepackage{siunitx}

\newcommand{\ie}{{\em i.e.,~}}

\title{\LARGE \bf
TSP-Bot: Robotic TSP Pen Art using High-DoF Manipulators
}

\author{Daeun Song, Eunjung Lim, Jiyoon Park, Minjung Jung and Young J. Kim
\thanks{The authors are with the department of computer science and engineering at Ewha womans university in Korea  {\tt\small \{daeunsong| ejunglim12|jiyoonpark13|juilejungh63\} @ewhain.net, kimy@ewha.ac.kr}}
}

\begin{document}

\maketitle
\thispagestyle{empty}
\pagestyle{empty}

\begin{abstract}

TSP art is an art form for drawing an image using piecewise-continuous line segments. We present TSP-Bot, a robotic pen drawing system capable of creating complicated TSP pen art on a planar surface using multiple colors. The system begins by converting a colored raster image into a set of points that represent the image's tone, which can be controlled by adjusting the point density. Next, the system finds a piecewise-continuous linear path that visits each point exactly once, which is equivalent to solving a Traveling Salesman Problem (TSP). The path is simplified with fewer points using bounded approximation and smoothed and optimized using B\'ezier spline curves with bounded curvature. 
Our robotic drawing system consisting of single or dual manipulators with fingered grippers and a mobile platform performs the drawing task by following the resulting complex and sophisticated path composed of thousands of TSP sites. As a result, our system can draw complicated and visually pleasing TSP pen art. 



\end{abstract}

\section{Introduction}

With the tremendous growth of digital technologies, digital art has become one of the largest art fields since the early 1960s. Early pioneers of digital art were not only artists but also engineers, computer scientists, and mathematicians who challenged traditional art standards with new technologies. 
Typically, most of the work focuses on investigating the production of artistic images in virtual space, which enables a wide variety of expressive and aesthetic styles using computer algorithms.  

Traveling Salesman Problem Art, abbreviated as TSP art, is one of the representative examples of creating artistic work using computer algorithms. It was first invented by mathematician Robert Bosh \cite{kaplan2005tsp}. 
TSP art is an art piece that represents the original digital image with piecewise-continuous line segments. 
TSP art involves not only the creative process of computer algorithms but also fits the nature of a robotic task, whose fundamental mission is to follow a path accurately and efficiently.

\begin{figure}[tb]
\centering
\includegraphics[width=0.75\linewidth]{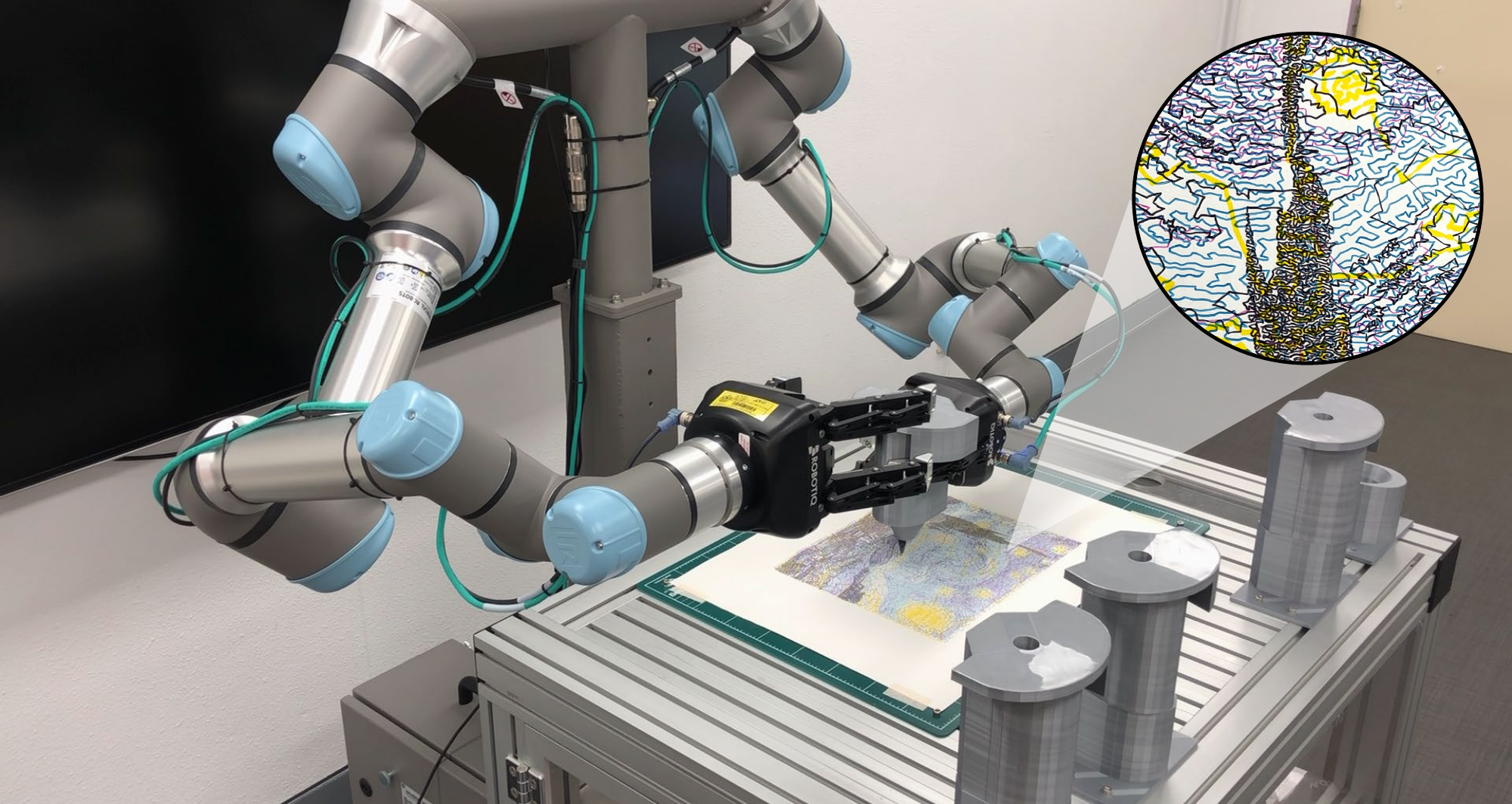}
\includegraphics[width=0.75\linewidth]{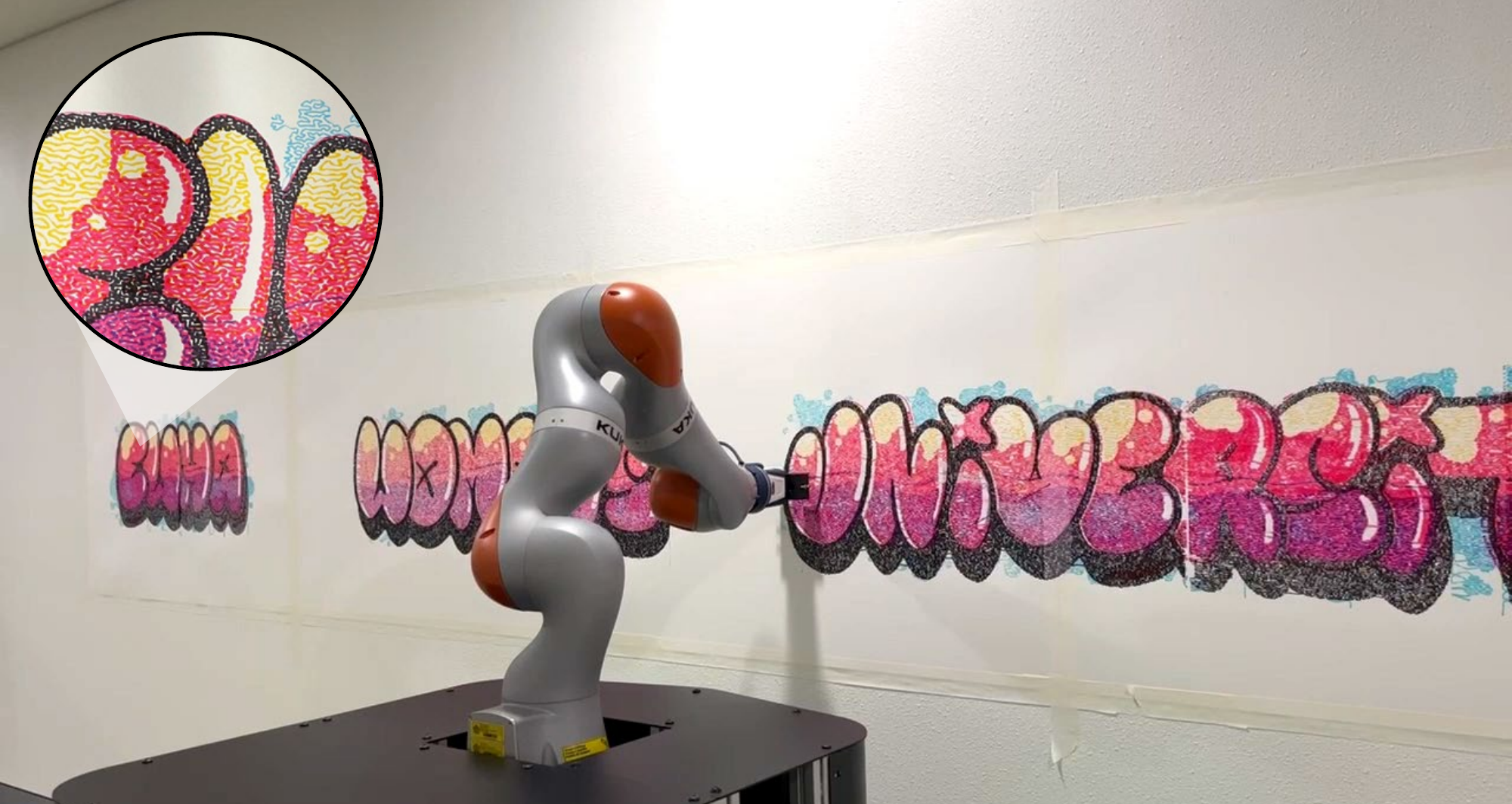} 
\caption{TSP-Bot system drawing TSP pen arts using a dual arm setup (top) and a mobile manipulator setup (bottom).}\vspace{-1.5em}
\label{fig:cover}
\end{figure}

As hardware technology advances, efforts have been made to bring these digitally generated artistic results into the physical space using machines \cite{cohen2016harold, lindemeier2015hardware}. As robots are capable of delivering long and complex motions,
we believe TSP art is best suited for the robotic drawing system, recognizing the original purpose of the robot. Our work focuses on a robotic TSP pen art system that is supported by complex and sophisticated motions. Our goal is not to supplant human artists but rather to aid and demonstrate the potential of interdisciplinary collaboration between robotics and art.
\newline \vspace{-0.5em}
\newline 
\noindent {\bf Main Results }
In this paper, we present a multi-color robotic pen drawing system, TSP-Bot, that transforms a digital raster image into long, continuous robotic paths that replicate the original image's tone and color and draw the result on a planar canvas surface (Fig.~\ref{fig:cover}). 
The system takes any raster image with color as input. In order to be reproduced by pens with a limited number of colors, the color image is {\em channel-split} into user-provided color spaces, such as the CMYK color space, and saved as separate image files. We use a stippling algorithm to displace points so that the points' density represents the image's tone. The system finds piecewise-continuous line segments that visit every point by solving TSP. We then perform path optimization with bounded curvature so the robot can follow smoothly. 
The drawing is rendered on a target canvas plane using our robotic hardware. 
We carry out drawing experiments using single and dual high degree-of-freedom (DoF) manipulators, the former with a mobile platform, to show that our system can create artistic and complicated TSP pen art in a physical space. We present diverse TSP drawing results. 

In summary, our technical contributions include:
\begin{itemize}
  \item A novel approach for processing color raster images suitable for limited-color-palette pens. This method involves splitting the input image into user-defined color spaces, such as CMYK, and using high-density TSP art to represent image tone accurately.
  \item A simple path optimization with bounded curvature to ensure smooth robot movement during the drawing process, coupled with the TSP solver.
  \item A novel drawing tool design with a tool-change mechanism to ensure robust pick and place using a 3-finger gripper. This new design enhances the reliability and versatility of the drawing system.
\end{itemize}

\section{Previous Work} \label{sec:prev}

\subsection{TSP art} \label{sec:TSPArt_prev}

TSP art was first presented in \cite{kaplan2005tsp}. It is an art form that reproduces an image's tonal quality with a single, continuous path by formulating the problem as a Traveling Salesman Problem (TSP). TSP art can be obtained by first placing some dots on the image and second connecting the dots with piecewise-continuous line segments. 
The stippling technique, to effectively reproduce the image's shading with the density of the points, has been explored by much research. \cite{deussen2000floating} used Lloyd's Voronoi-based optimization algorithm \cite{lloyd1982least} to create point patterns that look similar to human stipple artworks. \cite{secord2002weighted} enhanced the method to recreate faithful local stipple variations by adding weights to the centroid of the Voronoi region with regard to the density of the image. Most recently, \cite{Deussen:2017:WLS:3130800.3130819} combined the Linde-Buzo-Gray (LBG) algorithm~\cite{linde1980algorithm} with weighted Voronoi stippling~\cite{secord2002weighted}, dynamically splitting cells until achieving the desired number of representative vectors, and reformulated it to split and merge cells based on size, grayscale level, or image variance.

Once the stipples are placed, finding the shortest possible path that visits every stipple exactly once is equivalent to solving a TSP. Much research has been carried out to efficiently solve TSP, known as NP-Hard. Concorde \cite{campbell2007traveling} is an optimization solver widely regarded as the fastest TSP solver for large instances \cite{mulder2003million}. The resulting single line that resembles the original image's tone is considered an art form called TSP art. We observe that such complex and continuous paths are suitable for robotic drawing systems to reproduce them on a physical surface with its capability of sophisticated maneuverability. 




\subsection{Robotic Drawing}

The early history of creating drawing machines can be attributed to artistic work by Harold Cohens, AARON~\cite{aaron90}. 
AARON was a computer program capable of producing physical artwork using a plotting machine. With the recent advancement of robotic hardware, diverse artistic applications of robots have appeared. 
Paul the robot \cite{tresset13} is a robotic installation that creates portrait drawings by observing the target with the camera, mimicking the artist's stylistic signatures. eDavid \cite{lpd13} creates a painting using an industrial robot with the visual feedback and their Non-Photorealistic Rendering (NPR) algorithm. More research is being conducted using colors to produce more colorful and authentic drawings~\cite{karimov2023robot, scalera2024history}.
Recently, robotic drawing systems applying machine learning techniques have appeared. 
Integrating the algorithms for image segmentation and depth estimation for human-like stroke order planning was proposed~\cite{ilinkin2023stroke}. 
RoboCoDraw~\cite{wang2020robocodraw} is a personalized avatar character drawing system using a Generative Adversarial Network (GAN)-based style transfer approach. 
Additionally, researchers have explored the use of reinforcement learning to enable painting agents to learn optimal brush stroke placement \cite{schaldenbrand2020content}. 



The existing works focus on machine creativity and thus pay more attention to the painterly rendering algorithms, which still human outperform the machines. 
Recently, robotic pen drawing systems that focus more on the robot's capability have appeared. A system for drawing on non-planar surfaces using manipulator impedance control was introduced~\cite{songICRA18, songICRA19}. 
SSK~\cite{song2023ssk} further extends the system to draw on larger surfaces using a mobile manipulator by solving a coverage planning problem. 
Another exemplary instance highlighting the capabilities of robotic systems is a flexible and robust system adept at drawing on non-flat surfaces, prioritizing closed-loop planning for enhanced precision~\cite{liu2021robust}. 
Our work is also driven by the exploration of areas where machines excel compared to humans, all while ensuring the aesthetic integrity of the final outcome. 
Chitrakar~\cite{singhal2020chitrakar} proposes a robotic system that autonomously converts a human face image into a non-self-intersecting curve by solving TSP with a stippling method. 
We take a similar approach, but our system can draw more colorful drawings and is more suitable for autonomous robotic drawing with optimized drawing path and a pen-change mechanism.



\begin{figure*}[htb]
\centering
\includegraphics[width= 0.9\linewidth]{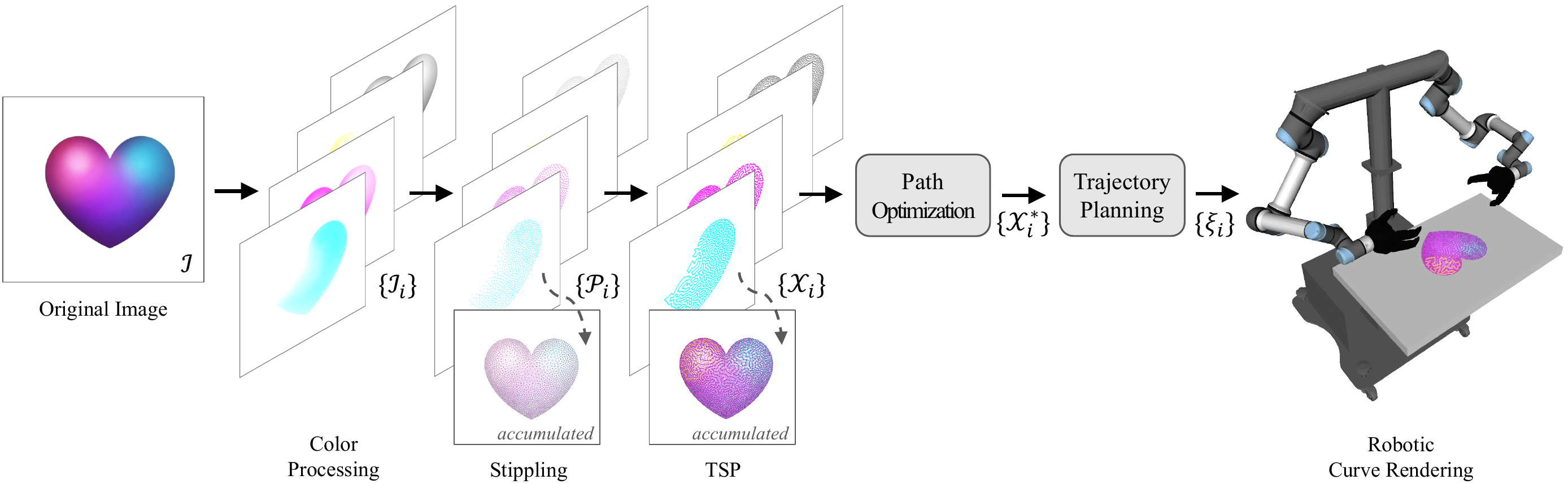} 
\caption{System overview. Given an original digital image, it is split into predefined color channels. A set of points is generated from each color channel using a stippling technique. We find a path that visits every point for each channel by solving a TSP on the generated points as TSP sites. The generated piecewise linear path is then smoothed and optimized with B\'ezier spline curves that ensure bounded curvature. We plan the robot's joint configurations so that the pen-holding end-effector follows the path for the drawing. 
} 
\label{fig:overview}
\end{figure*} 

\section{System Overview} \label{sec:overview}

We present the overview of our TSP robotic drawing system in Fig.~\ref{fig:overview}. Given a target image $\mathcal{I}$, in order to make it {\em drawable} for robots, we require a method to map it into the robot's configuration space. This paper employs TSP art to map the image to piecewise-continuous line segments within the canvas space. The path optimizer smoothens the trajectory by fitting cubic B\'ezier spline curves with bounded curvature. Subsequently, we determine the corresponding path in the configuration space by solving the path-wise Inverse Kinematics (IK) problem.

We generate TSP art using two main stages as follows:
\begin{enumerate}
    \item {\bf Stippling:} From $\mathcal{I}$, generate and place a set of points $\mathcal{P}\subset\mathbb{R}^2$ to replicate the tone of the original image.
    \item {\bf TSP Solving:} Find a cycling path $\mathcal{X}\subset\mathbb{R}^2$ that visits every point $\mathcal{P}$ once and returns to the first point.
\end{enumerate}
In order to reproduce the original image's color, we split the image $\mathcal{I}$ into predefined $n$ color channels $\mathcal{I}_i, i = 0,\cdots,n-1$, where $n$ is the number of colors. We repeat generating TSP art for each $\mathcal{I}_i$. 
In this paper, we follow a modern, four-color-process printing technique called CMYK~\cite{baqai2005digital} that splits the color space of the image into four color channels, cyan, magenta, yellow, and black. 
Afterward, we perform path optimization. We begin by applying the Ramer-Douglas-Peucker algorithm~\cite{ramer1972iterative} to simplify the piecewise linear paths with bounded Hausdorff distance between the original curve and the simplified curve. Then, we interpolate the linear paths using cubic B\'ezier spline curves $\mathcal{X}^{*} \subset\mathbb{R}^2$ with bounded curvature. 
We map the path to the end-effector's configuration $\mathcal{X}^{*} \subset\mathbb{R}^2 \rightarrow \tilde{\mathcal{X}}\subset SE(3)$ by projecting $\mathcal{X}^{*}$ onto the target 3D planar canvas space, with end-effector orientation perpendicular to the canvas. 

We finally find the joint configurations $\mathbf{\xi}$ corresponding to $\tilde{\mathcal{X}}$, which is fed into robots to perform drawing. Our system also considers the manipulator's reachability to decide the size of the drawable canvas space. We also designed a new pen-drawing tool for a 3-finger gripper to quickly and robustly switch between colored pens.

\section{Multi-color TSP Art} \label{sec:TSPArt}


In this section, we introduce our approach to map the input image $\mathcal{I}$ into a {\em drawable} robotic path $\mathcal{X}$. We take the TSP art idea with an additional color processing stage to reproduce the color likeness of the original image.

\subsection{Color Processing}

We begin by segmenting the color image into distinct color channels $\mathcal{I}_0, \cdots, \mathcal{I}_{n-1}$. Each channel is then independently processed to generate TSP art pieces. When reassembled, these TSP art fragments collectively replicate the color similarity of the original image. The method for channel separation can be controversial; we adopt the CMYK approach, commonly used for color printing, which separates images into cyan, magenta, yellow, and black channels. 
CMYK is effective due to its analogy to subtractive mixing, prevalent in both print and pen ink application. However, CMYK model may struggle to faithfully reproduce the original image color when working with fewer points. Further insights into alternative approaches are discussed in Sec.~\ref{sec:discussion}.


\subsection{Point Generation}

We perform stippling for each color channel individually, which involves transforming an image into a set of points $\mathcal{P} \subset \mathbb{R}^2$. This process results in desner point placement in darker regions and fewer points in brighter regions. Numerous stippling methods have been explored to achieve effective image reproduction. In this paper, we adopt the LBG Stippling method~\cite{Deussen:2017:WLS:3130800.3130819} which combines a variant of the Linde-Buzo-Gray algorithm \cite{joshi2014image} with weighted Voronoi stippling \cite{secord2002weighted}. 
The method progressively divides Voronoi cells from a single point until the desired point density, determined by the weight function, is reached. Additionally, the method allows the merging of the neighboring cells once the point density becomes excessive. 

\subsection{TSP Path Generation} 
The generated points $\mathcal{P}$ can already form an art piece, feasible for robotic motion. However, repetitive up-and-down motions are time-intensive. To address this, we connect all the points in $\mathcal{P}$ to establish a singular path $\mathcal{X}(t) \subset \mathbb{R}^2$, which makes the robot easier to follow. Finding this path is equivalent to solving TSP, a widely known NP-hard problem. Significant research has been directed towards efficient approximated solutions for TSP. In this work, we utilize the Concorde solver, which relaxes the problem into a Linear Program (LP) and iteratively fixes potential fractional solutions through a cutting plane algorithm~\cite{dantzig1954solution}. Using the Lin–Kernighan heuristic~\cite{karapetyan2011lin}, which iteratively improves a tour by exchanging pairs of edges while utilizing edge removal, insertion, and recombination to minimize edge crossings, we can generate a path without any edge crossings.



\section{Robotic Drawing} \label{sec:robotic}

This section outlines the optimization of the drawing path and describes selecting the appropriate canvas dimensions based on the manipulator's reachability and relocating the drawing end-effector pose accordingly. Furthermore, we detail the robotic curve rendering method, which involves determining joint configurations to execute the drawing task.



\subsection{Path Optimization}

Tracing a TSP path composed of piecewise linear segments $\mathcal{X}(t)$ that interpolate $\mathcal{P}$ is feasible for an end-effector to follow, but it is not optimal for realizing the robot motion. We optimize the path $\mathcal{X}^{*}(t)$ to be better suitable for robotic tracing while approximating the original path with bounds on distance and curvature. The Ramer-Douglas-Peucker algorithm simplifies the linear path by decimating some sub-path that does not significantly contribute to the curve's shape. This process involves recursively subdividing the path, checking if the distance between a sub-path and the original path is below a threshold $d_\epsilon$, and then discarding it.

\begin{figure}[h]
\centering
\includegraphics[width=0.52\linewidth]{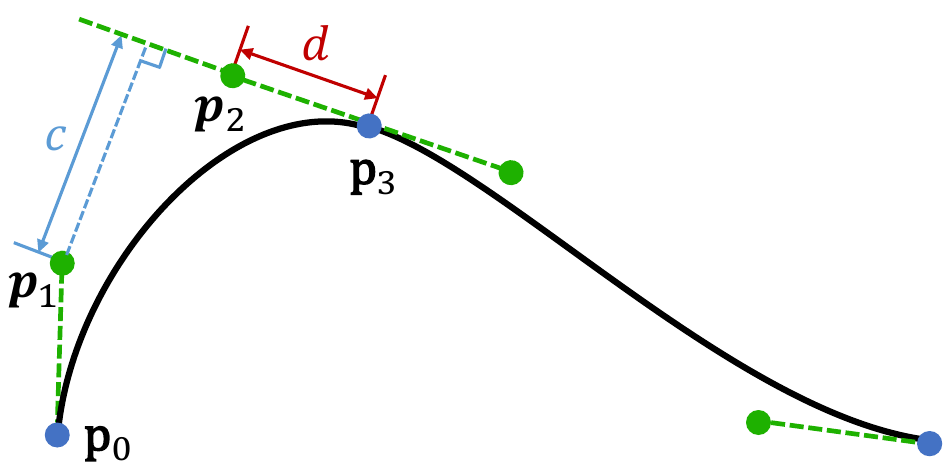}\vspace{-0.5em}
\caption{Interpolating cubic B\'ezier curve. $\mathbf{p}_i$'s represent the control points for the first B\'ezier spline that interpolates $\mathbf{p}_0$, $\mathbf{p}_3$.}\vspace{-0.5em}
\label{fig:bezier}
\end{figure}

After acquiring a simplified piecewise linear path that interpolates a reduced number of points $\mathcal{P}^{*}$, we interpolate them using cubic B\'ezier spline curves. The curvature of these curves is bounded by $\kappa_\epsilon$, determined by the maximum acceleration of the end-effector. The entire path comprises $|\mathcal{P}^{*}|-1$ spline curves, each with four control points. Fig.~\ref{fig:bezier} illustrates two spline curves connected at $\mathbf{p}_3$. The curvature of the B\'ezier spline curve at $\mathbf{p}_3$ is evaluated as:
\begin{equation}\label{eq:bezier}
\begin{aligned}
   \kappa = \frac{2d}{3c^2} \,,
\end{aligned}
\end{equation}
where $d=||\mathbf{p}_2-\mathbf{p}_3||$ and $c$ represents the distance between $\mathbf{p}_1$ and the line formed by $\mathbf{p}_2$ and $\mathbf{p}_3$ \cite{farin2014curves}. 
Thus, $\kappa$ can be controlled by fixing $c$ and increasing $d$ by displacing $\mathbf{p}_2$ along $\mathbf{p}_2-\mathbf{p}_3$ with some scaling factor $s>0$. The final path $\mathcal{X}^{*}$ is obtained by minimizing $s$ such that $s\kappa_i > \kappa_\epsilon, \forall i \in |\mathcal{P}^{*}|-1$ where $\kappa_i$ denotes the curvature for $i$th spline.


Optimizing a path of more than 50,000 control points may require a significant amount of computation time. Thus, we employed rather simple path optimization and smoothing techniques to save time. Moreover, since the preceding TSP solver already produced spatially coherent and optimal drawing routes, our simple approach still produces a highly optimal path and prevents abrupt robot motion.

\subsection{Maximum Size of Canvas Space} \label{sec:canvas}

\begin{figure}[tb]
\centering
\subfigure[Dual Manipulators]{\includegraphics[height=5.0cm]{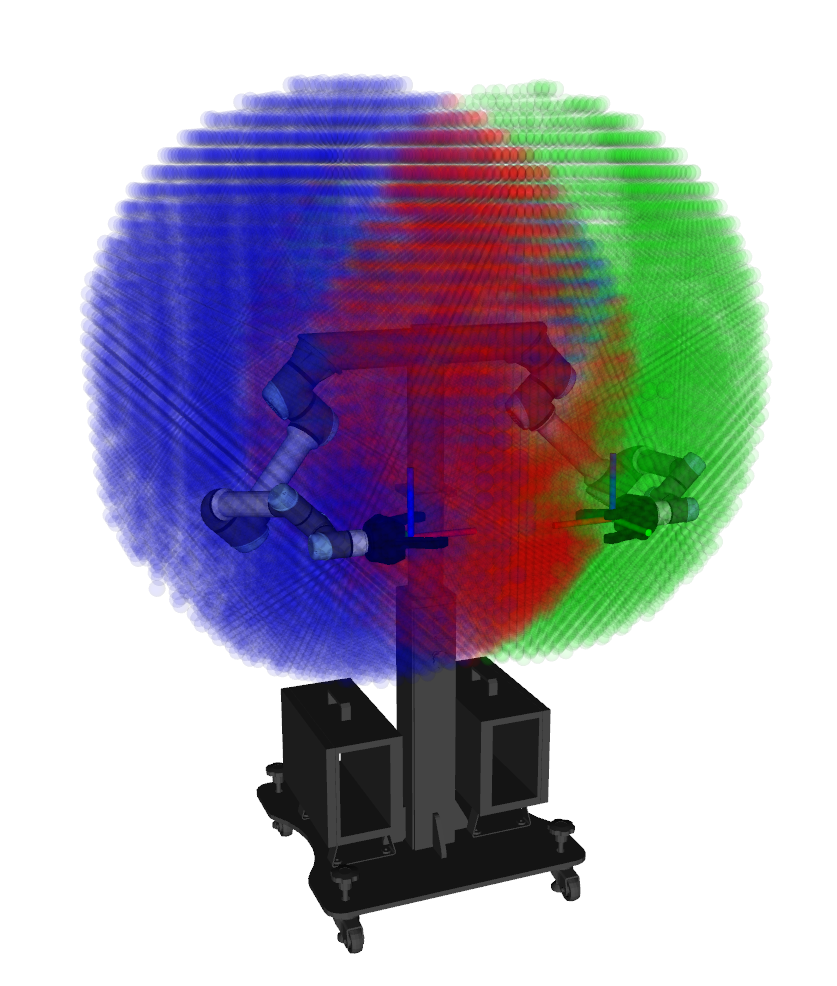}}
\subfigure[Mobile Manipulator]{\includegraphics[height=4.9cm]{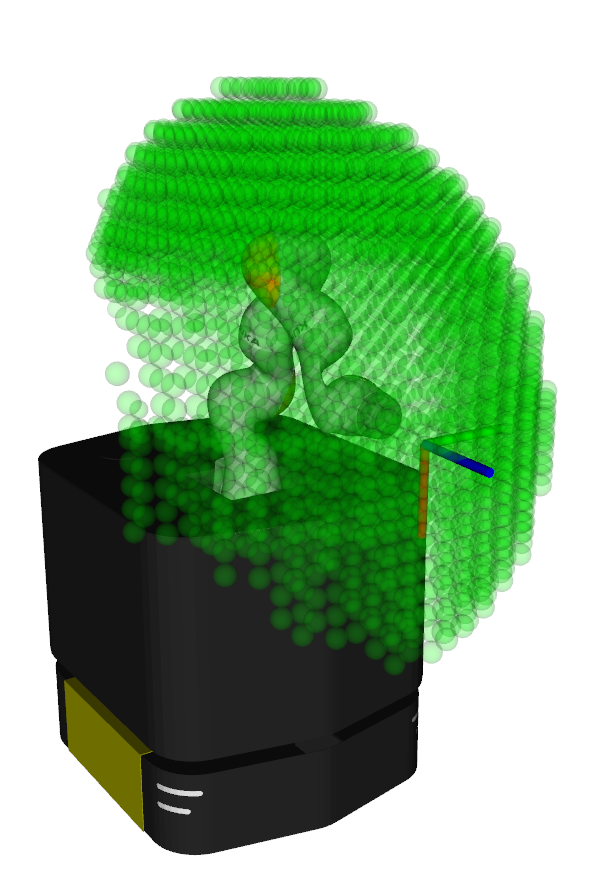}}
\caption{Reachability of the manipulators. Green or blue spheres in (a) and (b) represent the points that are reachable by each manipulator with a fixed end-effector orientation. Red spheres in (a) represent the region that is reachable by both manipulators.}\vspace{-1.0em}
\label{fig:reachability}
\end{figure}

After generating the final 2D drawing path, the next step before executing the robotic task is to map the path onto the 3D real-world space. We determine the maximum size of the drawing canvas based on the robot's reachability. The reachability of the robot is defined by discretizing the robot's Cartesian workspace and solving the inverse kinematics problem for each discrete point to check if it is reachable~\cite{porges2015reachability}. The reachable points are saved as shown in Fig.~\ref{fig:reachability} with colored spheres. The intersection between the reachable point set and the target surface is used to determine the canvas dimension. For dual-arm, we separate the drawing task by color channels and find the intersection between the reachable point set by both arms and the target surface. For mobile manipulator, we split the canvas into sub-canvas based on the manipulator's drawing size. Then, we repeat the process of drawing and shifting to the next sub-canvas to complete the drawing. After determining the canvas space, we obtain the target drawing poses by projecting the drawing path onto the planar canvas space with fixed end-effector orientation in the opposite direction of the surface normal: $\mathcal{X}^{*} \subset\mathbb{R}^2 \rightarrow \tilde{\mathcal{X}}\subset SE(3)$. 


\subsection{Robotic Curve Rendering}

Once the target drawing poses $\tilde{\mathcal{X}}$ are decided, they are fed to robots. Robotic drawing is equivalent to finding a continuous path in the robot's configuration space so that its end-effector follows the given path; this problem is also known as path-wise IK~\cite{chitta2012moveit,beeson2015trac}. Moreover, path-wise IK is suitable for setting a robot's kinematic and dynamic constraints while following the robot trajectory. In the case of robot drawing, achieving a feasible motion with no sudden jumps is crucial. Plus, it is essential to follow the given end-effector poses accurately not to ruin the resulting drawing. Thus, we solve the path-wise IK problem by iteratively solving the IK for the end-effector poses $\tilde{\mathcal{X}}$ with the minimum distance objective in the configuration space.



\begin{figure*}[t]
\centering
\includegraphics[height=1.4in]{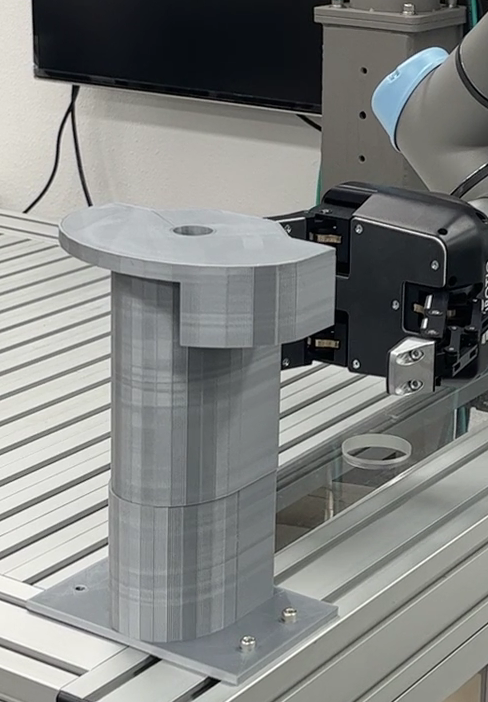}
\includegraphics[height=1.4in]{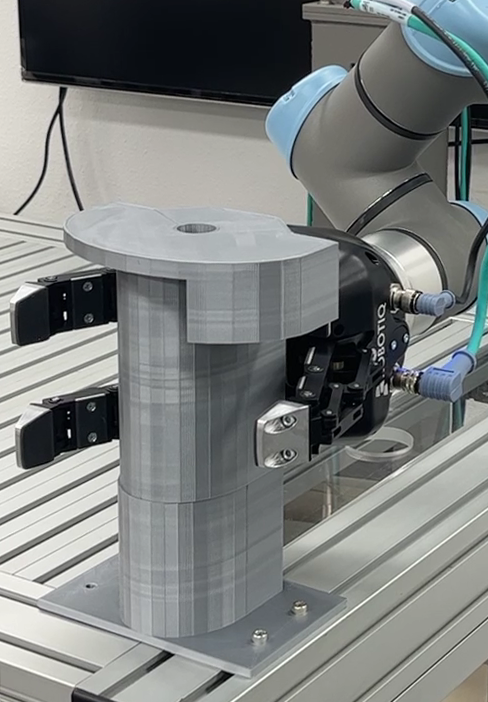}
\includegraphics[height=1.4in]{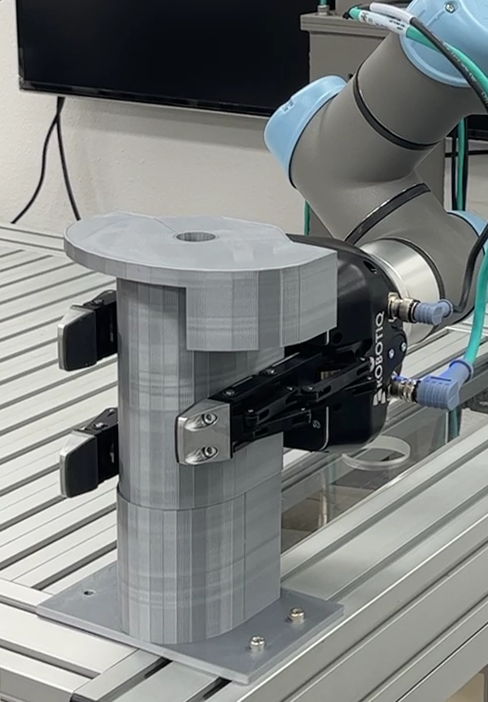}
\includegraphics[height=1.4in]{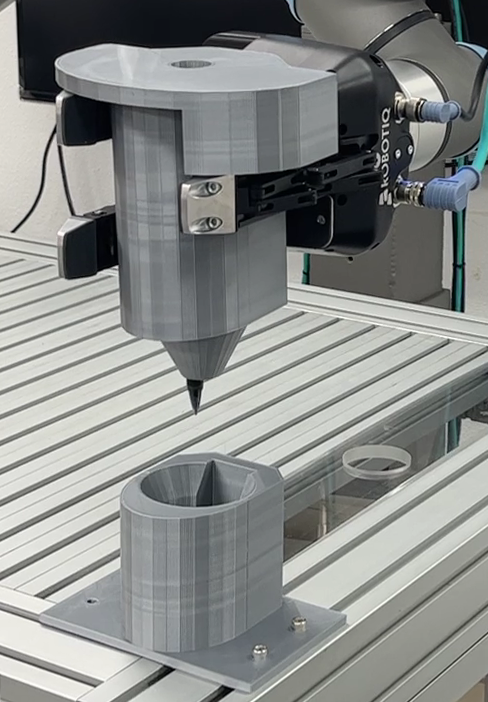}
\includegraphics[height=1.4in]{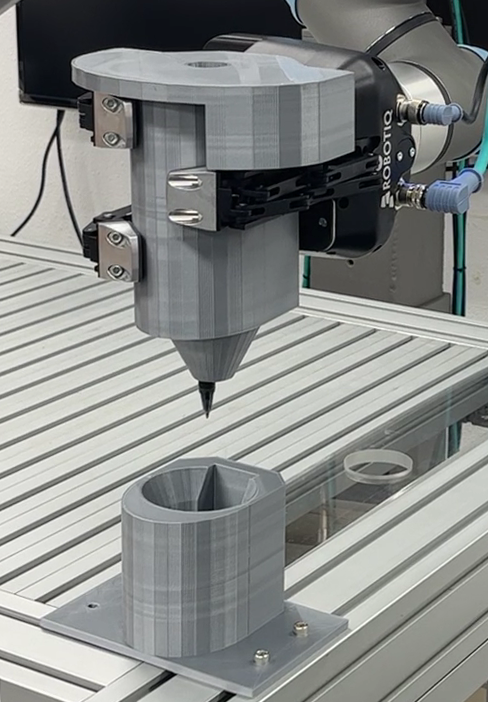}
\caption{Pen tool picking sequences using a 3-finger gripper in five stages: from left to right, ready pose, approach, half grasp, vertical lift, and full grasp. Placing the tool back to the docking structure is in reverse.} \vspace{-0.5em}
\label{fig:tool}
\end{figure*}

\section{Drawing Tool}

Robotic pen drawing requires firmly holding the pen to the robot's end-effector as the task needs to resist contact force from the canvas surface. A potential solution for this requirement involves rigidly attaching the pen to the robot, which makes human intervention inevitable for changing the pen to a different color. Alternatively, one can employ a robotic gripper to grasp and exchange the pen when needed. This mechanism essentially corresponds to a pick-and-place task, which is particularly challenging due to various uncertainties of sensing and dynamics in the real world. This section elaborates our approach to the pen-changing problem ensuring robustness for multi-color drawing systems.



Fig.~\ref{fig:tool} shows snapshots of tool-change sequences using the pen-holding tool and its docking structure. We design a pen-holding tool that the 3-finger gripper can robustly grasp even under slight motion perturbations due to position or motion errors, which also ensures that the gripper always holds the pen in the same pose. We flatten the side of the tool so that when the gripper is half-closed, it can still adjust the vertical orientation by the fingers. We created upper plates in the holding structure that match the fingers' height so that they can fine-tune the gripper's horizontal orientation when lifted vertically by a half-closed gripper. The tool shape that matches the palm of the gripper once again guarantees to grasp the tool when closing the fingers robustly. 
The pen tool's docking structure is a rounded concave shape that matches the surface contour of the pen-holding tool. It allows the tool to slide into the bottom when released from the gripper. Another critical and practical aspect of the docking structure is preventing the pen from drying during the long drawing session. Our drawing tool and its docking structure design are experimentally proven robust for many tool changes.


\begin{figure}[htb]
\centering
\subfigure[Dual manipulators and the pen-holding tools]{\includegraphics[height=1.45in]{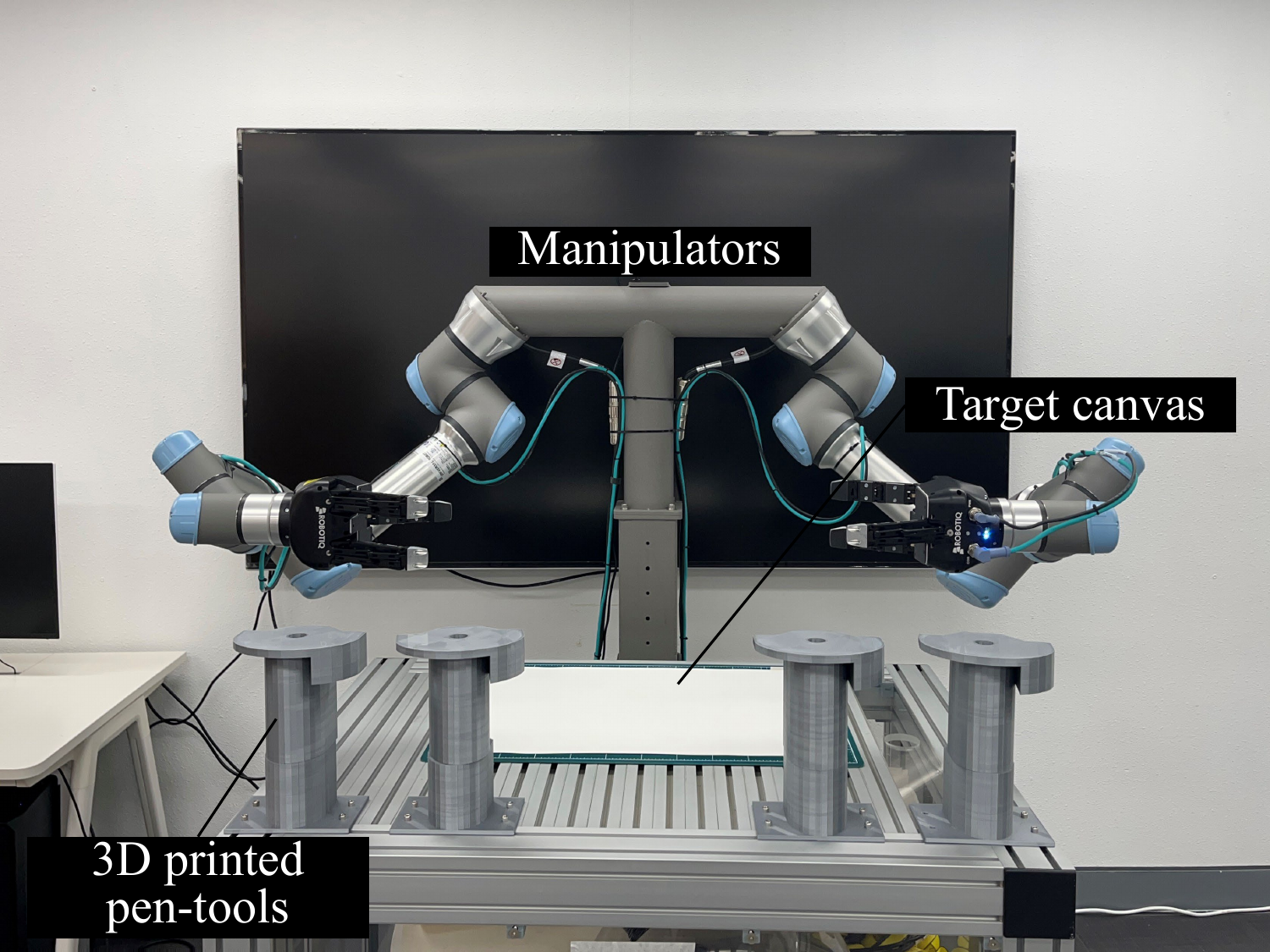}\hspace{0.2em}\includegraphics[height=1.45in]{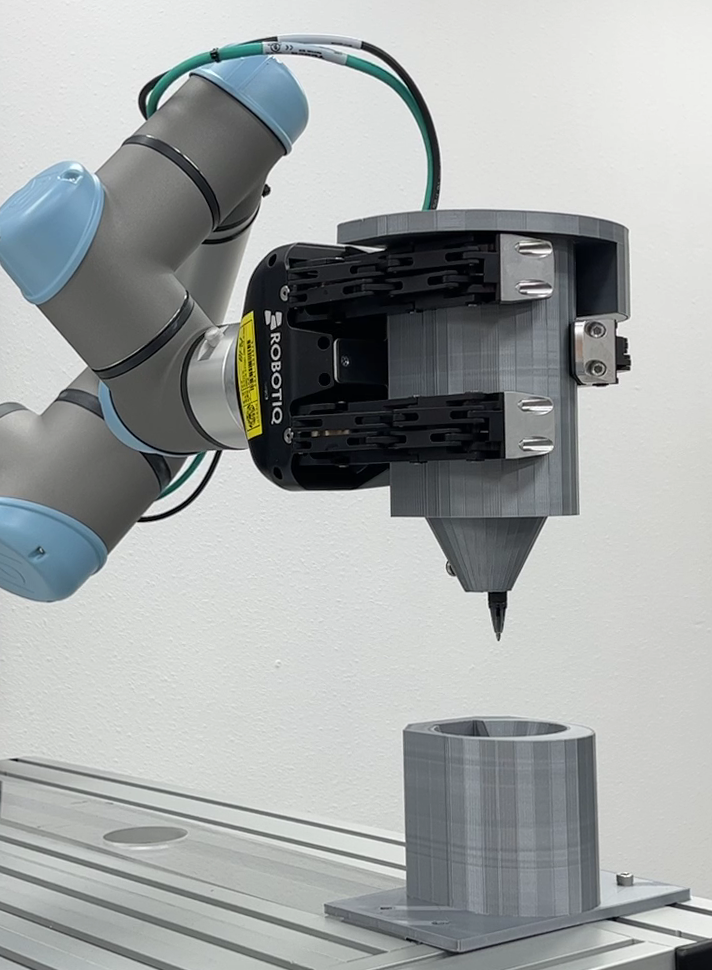}
}\label{fig:ur}
\subfigure[Mobile manipulator and the pen-holding tool]{\includegraphics[height=1.45in]{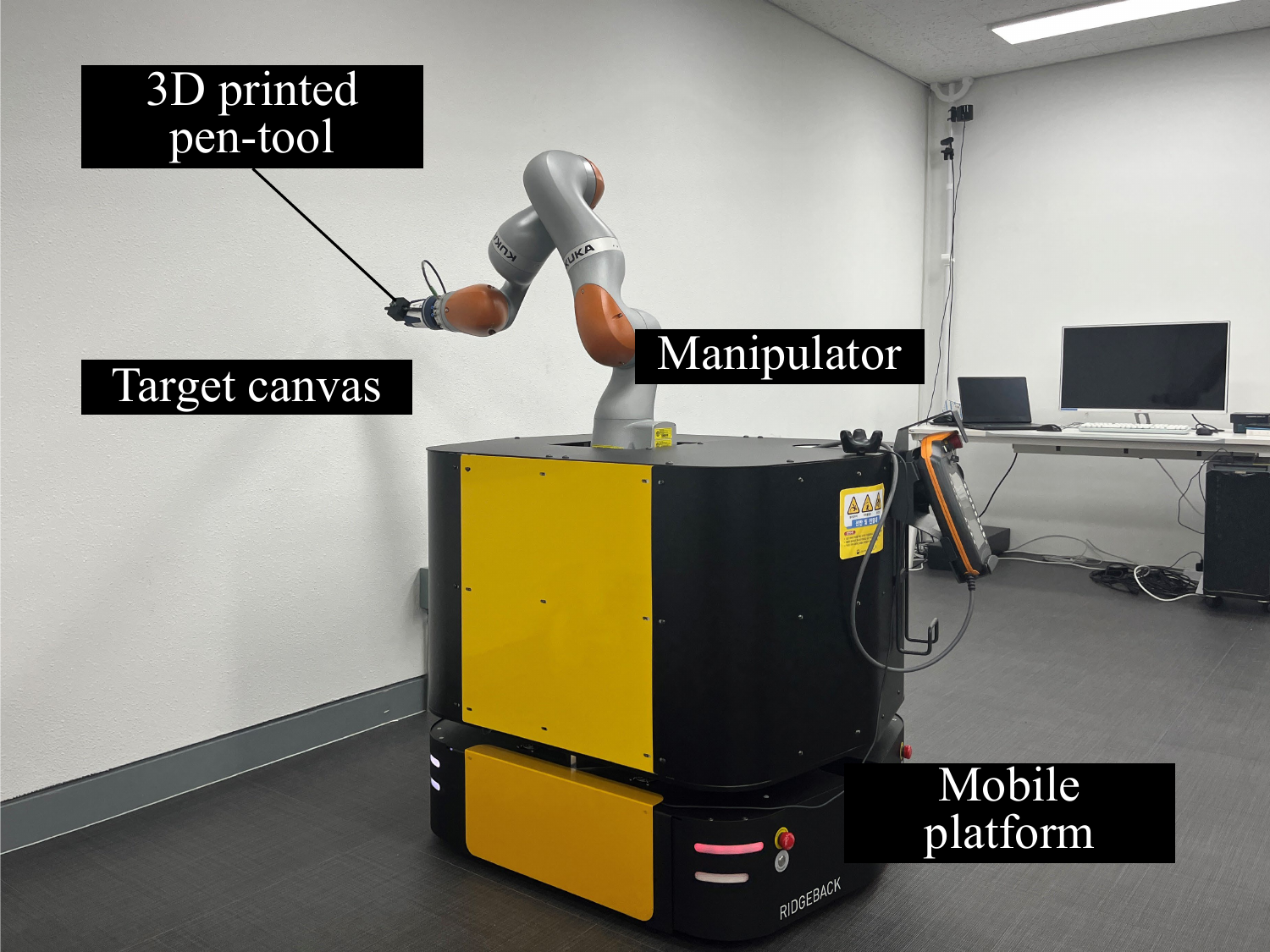}\hspace{0.2em}\includegraphics[height=1.45in]{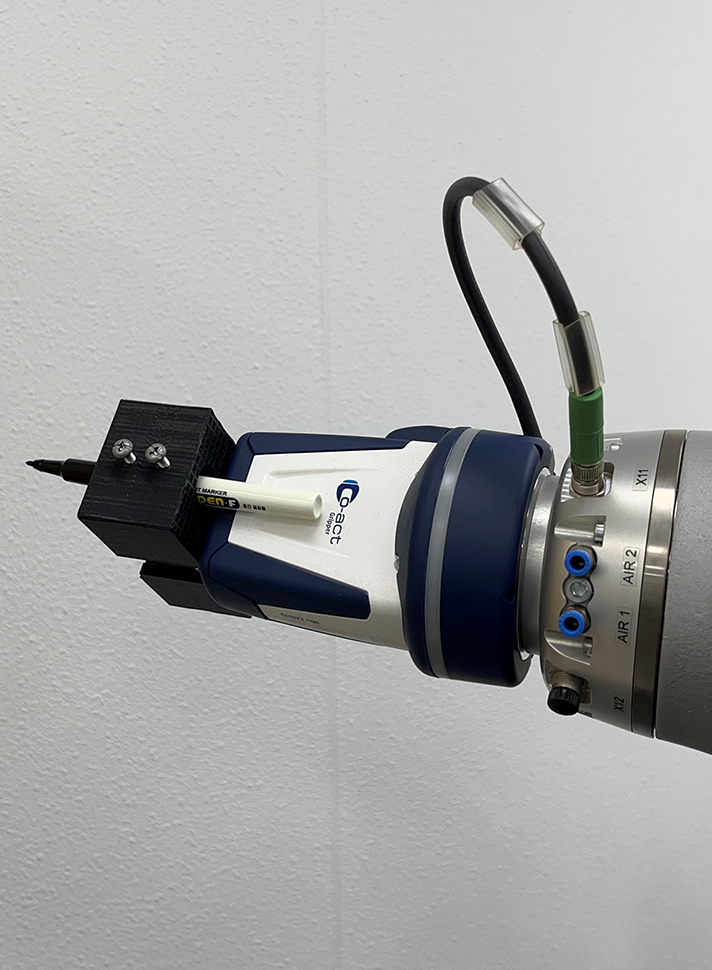}}\label{fig:iiwa}
\vspace{-0.5em}\caption{Robotic TSP pen-art drawing system setup} \vspace{-0.5em}
\label{fig:setup}
\end{figure}

\section{Results and Discussions} \label{sec:result}

In this section, we present our experiment setting and discuss the results. In particular, we show our drawing results with two different color-processing methods performed with two robotic hardware setups, one with a mobile manipulator and another with dual manipulators.


\subsection{Implementation Details}

As shown in Fig.~\ref{fig:setup}, we implemented our robotic drawing system with two different robotic hardware setups:
\begin{itemize}
    \item {\bf Dual Drawing Manipulators Setup} consists of two UR5e manipulators, each equipped with a Robotiq 3-finger adaptive gripper. We designed a new pen tool that allows our gripper to hold the pen firmly despite sensing and mechanical errors. The drawing process is fully automated thanks to the gripper and our pen tool change mechanism. 
    \item  {\bf Mobile Drawing Manipulator Setup} consists of a KUKA LBR iiwa 7 R800 manipulator mounted on top of the omnidirectional mobile platform Ridgeback from Clearpath Robotics. A 3D-printed pen tool is rigidly attached to the robot's end-effector, which requires a manual change of colors. Thanks to its mobility, unlike previous robotic drawing systems, it is not limited to the drawing canvas size. 
\end{itemize}

We use Robot Operating System (ROS) Melodic framework under Ubuntu 18.04 operating system to communicate with the robots and perform the drawing task. 
We use C++ and Python for programming, which runs on a PC equipped with Intel i7 CPU and 32 GB memory. We use MoveIt!~\cite{chitta2012moveit} with TRAC-IK~\cite{beeson2015trac} inverse kinematics solver for computing the robot trajectory following the Cartesian path. 

\begin{table*}[htb] 
\centering
\caption{Robotic Drawing Experimental Statistics}
\begin{tabular}{rcccccccc} 
 \toprule
 \textbf{Robotic Hardware} &  \multicolumn{2}{c}{\textbf{Dual Arm (Fig.~\ref{fig:result_dual})}} & \multicolumn{3}{c}{\textbf{Mobile Manipulator (Figs.~\ref{fig:cover},~\ref{fig:result_mm})}} \\
 \cmidrule(lr{0.5em}){2-3} \cmidrule(lr{0.5em}){4-6} 
 \textbf{Drawing}&  \hspace{0.1cm}Starry Night  & Big Ben\hspace{0.1cm} &\hspace{0.1cm} Heart & Violet$^{*}$ & EWU\hspace{0.1cm}\\
 \midrule
    Canvas Size ($\unit{mm}^2$) &   $\numproduct{315 x 250}$ &  $\numproduct{214 x 300}$ & $\numproduct{400 x 350}$ & $\numproduct{850 x 300}$ & $\numproduct{3600 x 400}$   \\
    \# of Stippled Points &  81,591  & 76,257 & 21,664 & 95,155 & 154,475  \\ 
Stippling Time (sec.) &  32 & 34 &  2 & 35 & 32  \\ 
    \hspace{0.1cm}TSP Solving Time (sec.) & 6 & 11 & 4 & 17 & 43  \\ 
    Drawing Time (min.) & 124 & 63 & 61  & 585  &  662 \\ 
 \bottomrule
\end{tabular}
\label{table:stat}
\end{table*} 

\begin{figure*}[htb]
\centering
\subfigure[Starry Night]{\includegraphics[height=1.35in]{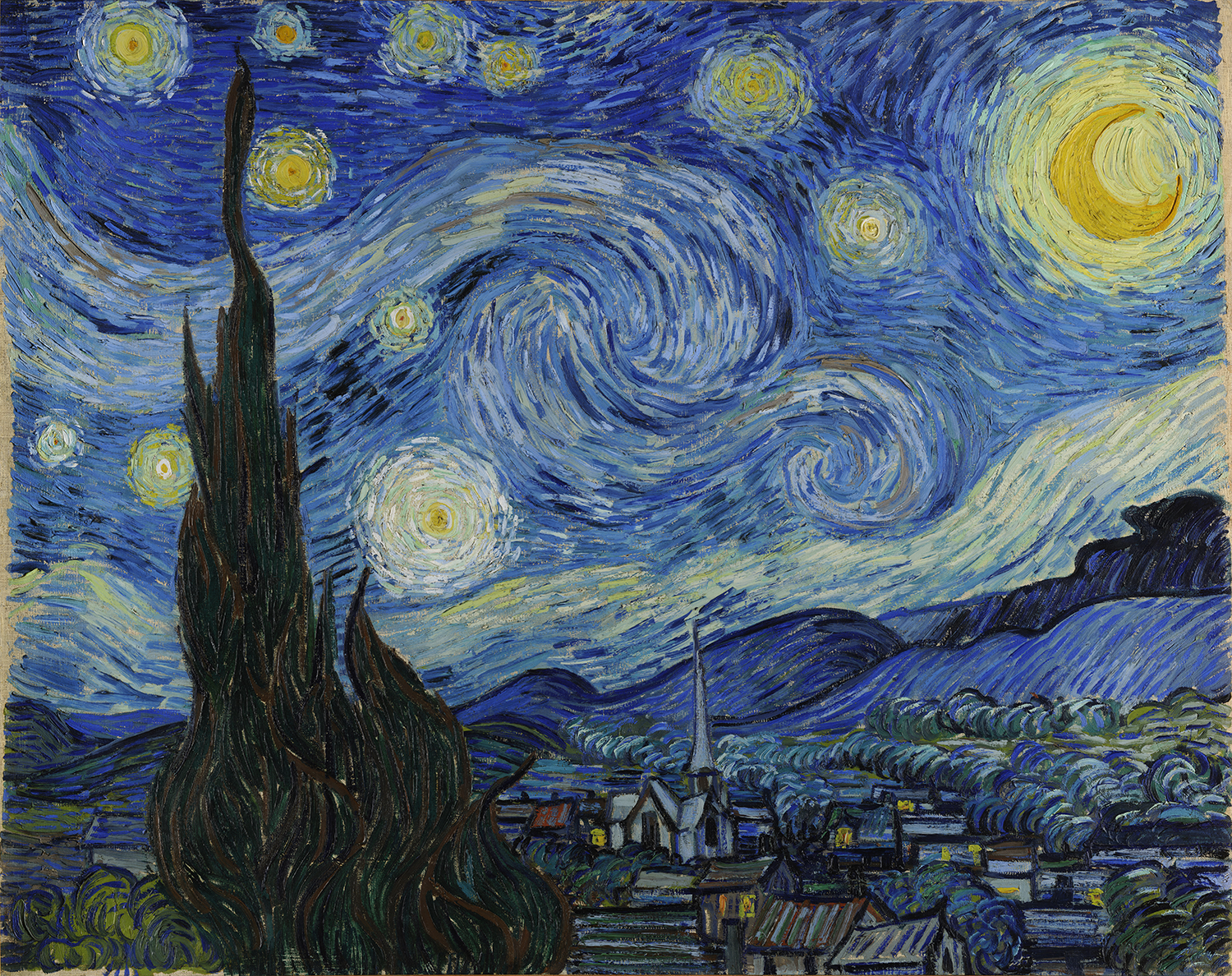} \hspace{0.1em}
\includegraphics[height=1.35in]{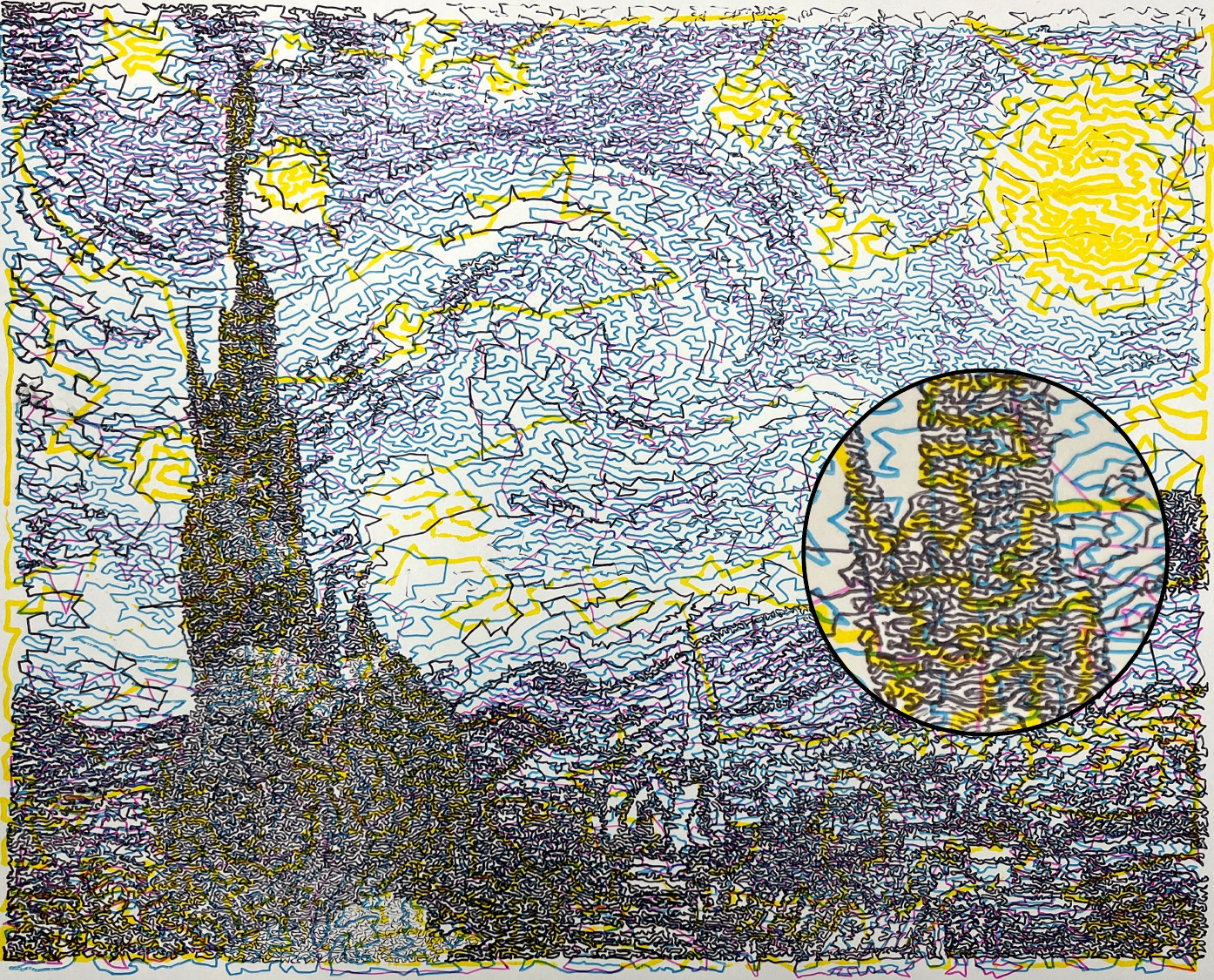}
}
\subfigure[Big Ben]{\includegraphics[height=1.35in]{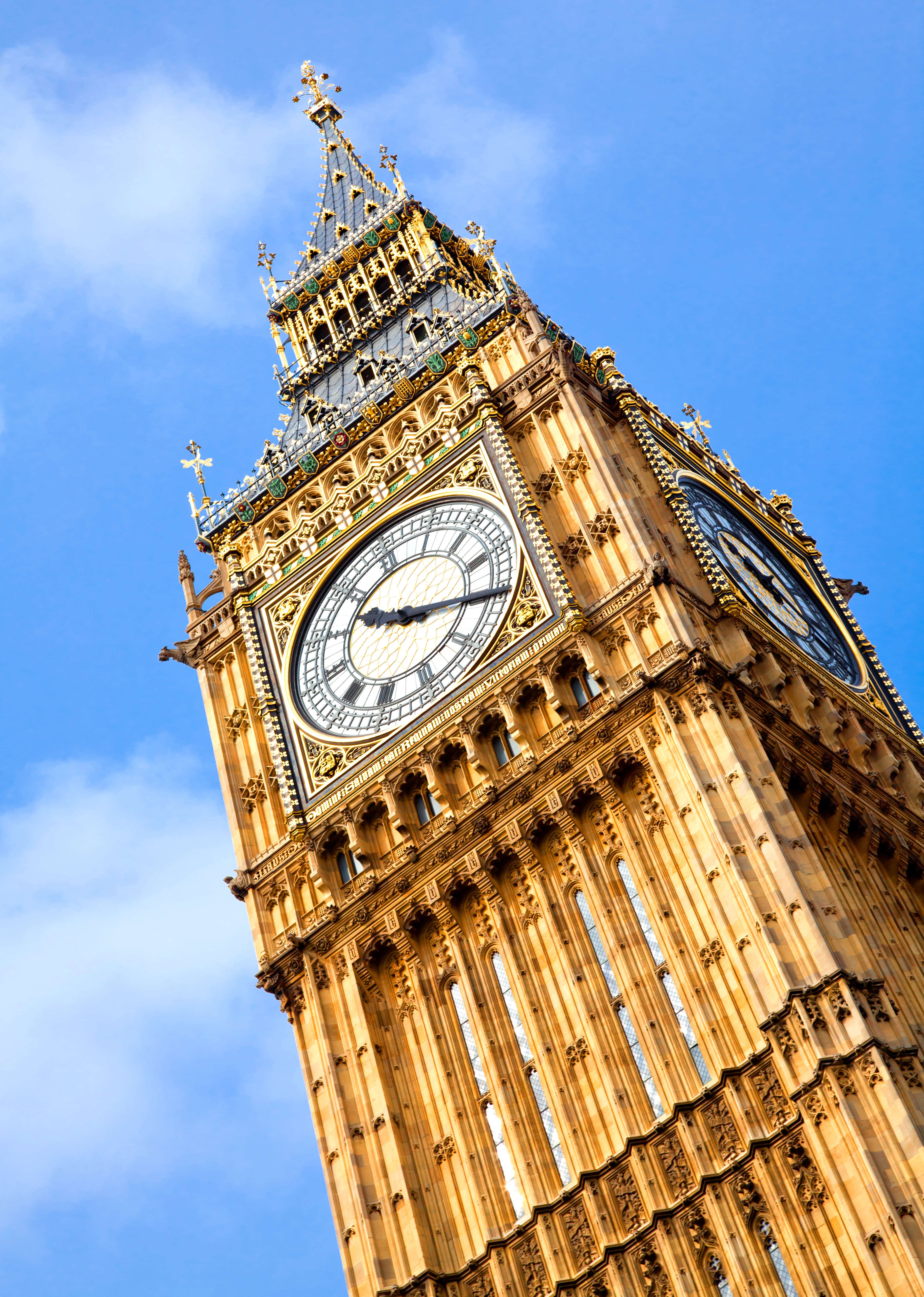}\hspace{0.1em}
\includegraphics[height=1.35in]{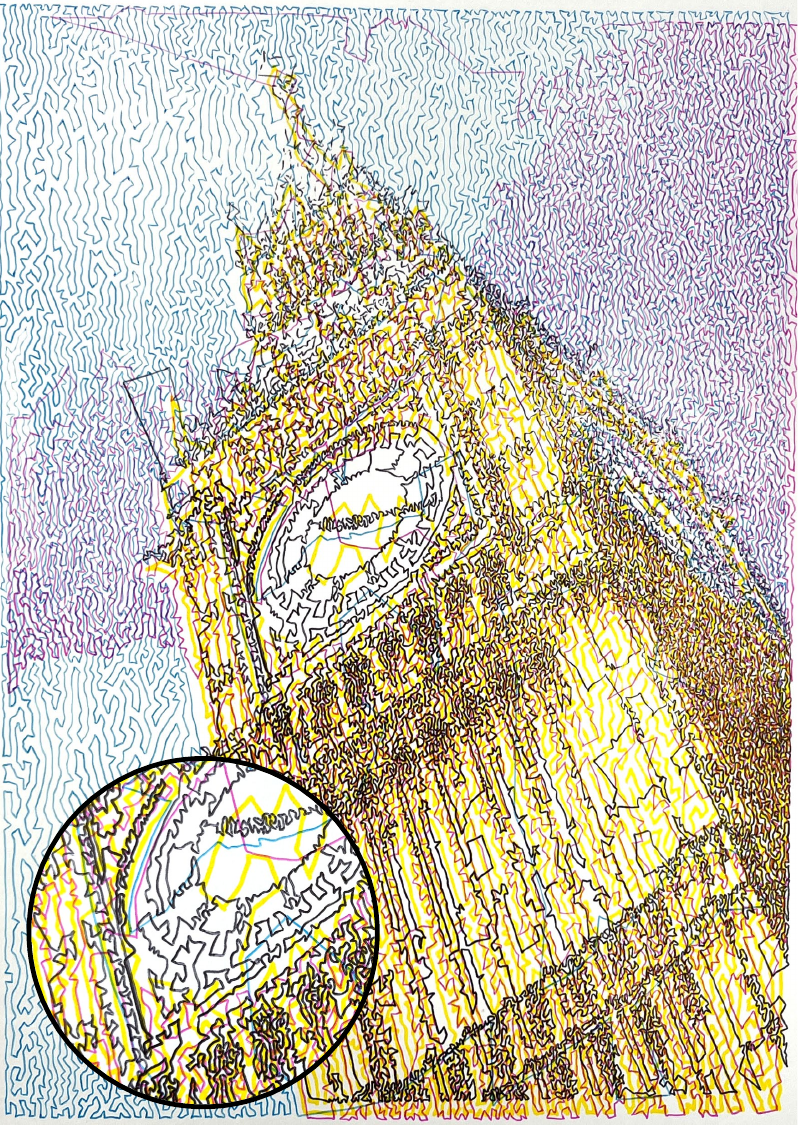}}
\vspace{-0.5em}\caption{TSP pen art results produced by the dual manipulators (left: input image, right: TSP drawing)} \vspace{-1.0em}
\label{fig:result_dual}
\end{figure*} 

\subsection{Drawing Results}

\begin{figure}[htb]
\centering
\subfigure[Heart (left: input image, right: TSP drawing)]{\includegraphics[height=2.6cm]{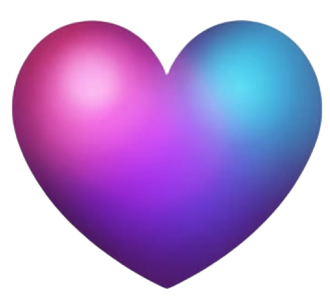} \includegraphics[height=2.5cm]{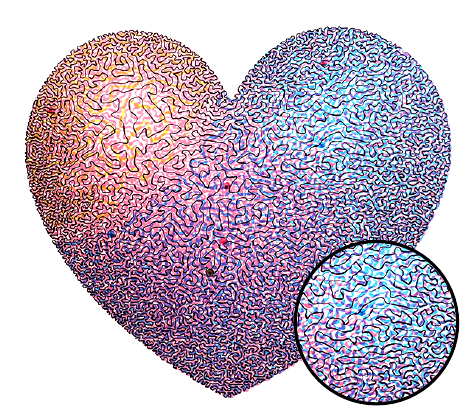}}
\includegraphics[height=2.6cm]{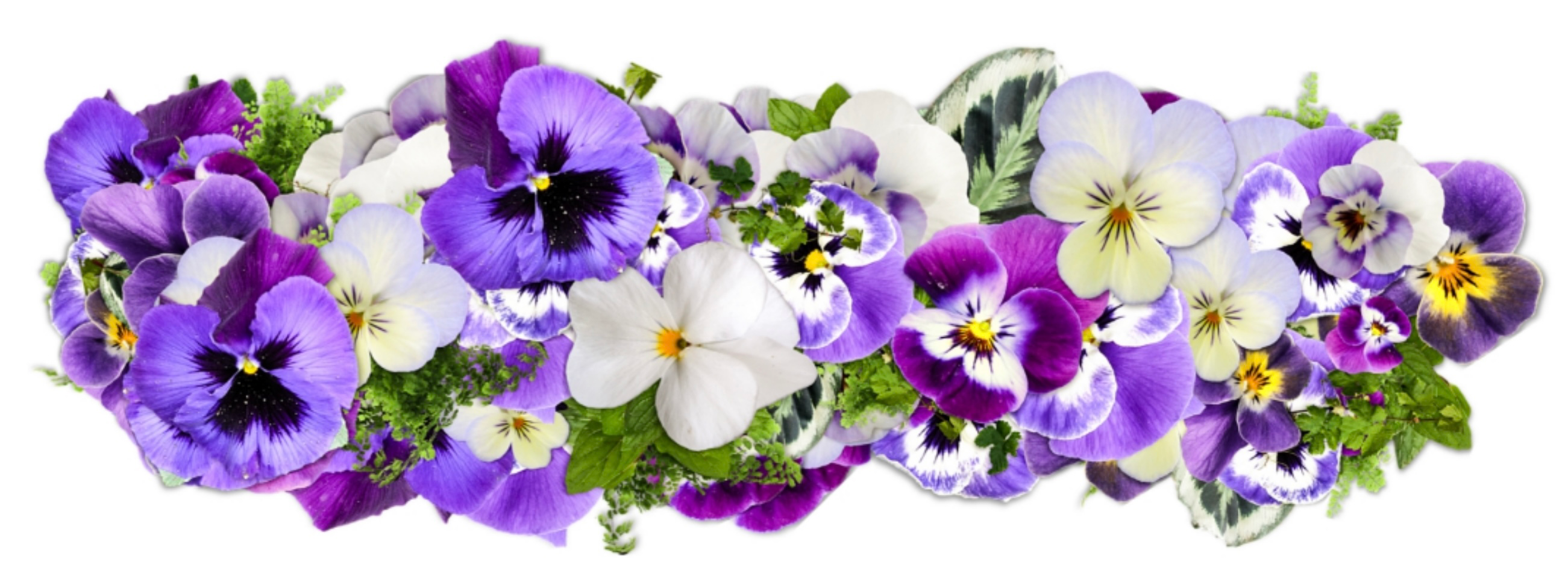} 
\subfigure[Violet (top: input image, bottom: TSP drawing)]{
\includegraphics[height=2.55cm]{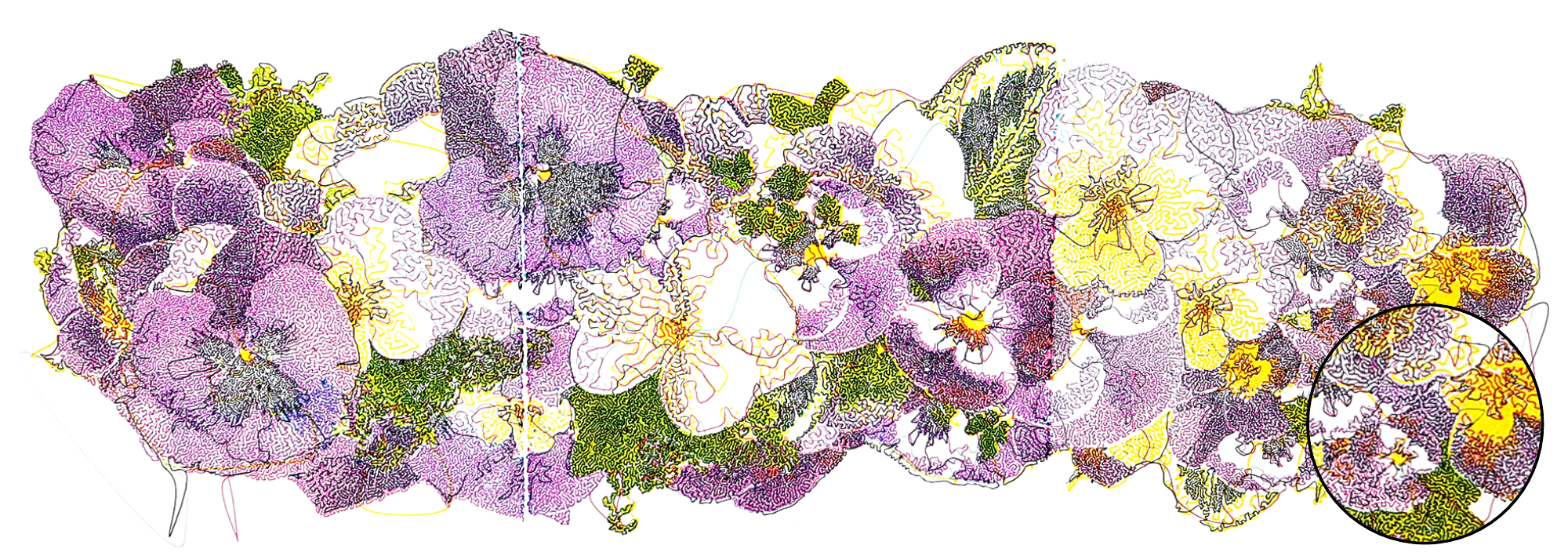}}
\caption{TSP pen art results produced by the mobile manipulator}\vspace{-1.5em}
\label{fig:result_mm}
\end{figure}

The statistics for the experimental drawings shown in Figs.~\ref{fig:cover},~\ref{fig:result_dual} and \ref{fig:result_mm} are provided in Table~\ref{table:stat}. 
These statistics include the size of the drawing and the number of stippled points used to solve TSP. Additionally, the time consumed to run the stippling algorithm, solve TSP, and execute the robot is presented. The number of points and the times in the statistics represent the sum of values for every color. Please note that the drawing result of Violet in Fig.~\ref{fig:result_mm}, marked with * in Table~\ref{table:stat}, is generated with Weighted Voronoi stippling algorithm~\cite{secord2002weighted} as a stippling method. 
Regarding TSP solving time, we impose a time-bound to ensure that the data generation process does not take an excessive amount of time. 
During the drawing process, we operate the robot at a low speed, specifically 20\% of the maximum joint velocity. This is to ensure the safety of both the robot and any collaborators, such as humans. 

\begin{figure}[htb]
\centering
\subfigure[Original TSP path]{\includegraphics[width=0.35\linewidth]{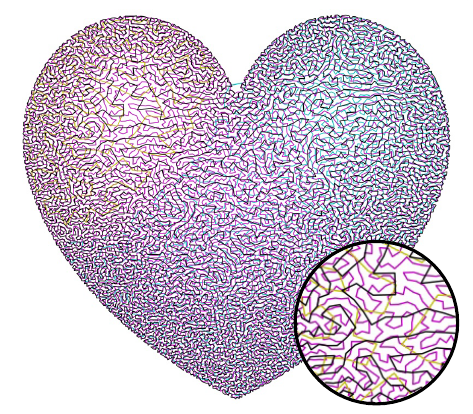}}\hspace{0.5em}
\subfigure[Optimized path]{\includegraphics[width=0.35\linewidth]{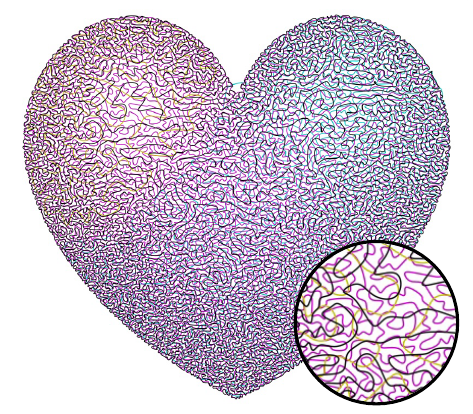}}
\caption{Digital drawing result of Heart before and after the path optimization}\vspace{-1.0em}
\label{fig:opt}
\end{figure}

Fig.~\ref{fig:opt} shows the digital drawing result of the Heart before and after the path optimization. 
Note that the TSP path in Fig.~\ref{fig:opt}(a) without path optimization produces a similar result like Chitrakar~\cite{singhal2020chitrakar}. 
The number of points decreased from 21,664 to 16,720, which means about 15\% of the points were removed while the shape of the result is non-distinguishable between the two. 
We experimentally set the parameters of $d_\epsilon$ and $\kappa_\epsilon$ to 0.5 and 2.0, respectively, given the simplification rate of 15\% and the bound for maximum end-effector acceleration. 

Fig.~\ref{fig:result_dual} shows the robotic pen drawing results drawn with a dual-arm robotic setup. 
Fig.~\ref{fig:cover} and Fig.~\ref{fig:result_mm} show the robotic pen drawing results drawn with the mobile manipulator setup. The Heart is drawn in one place with no mobile platform moving. The Violet drawing was split into three regions, and the Ewha Womans University (EWU) graffiti was split into nine regions according to the appropriate canvas space measured in Sec.~\ref{sec:canvas}. 

\subsection{Discussions} \label{sec:discussion}

{\bf Color Splitting Methods:}
The CMYK color channels we use for the system are widely used for printing materials that combine colors. This is suitable for our robotic pen drawing system, which also uses a limited number of colors and aims to achieve a similar color to the original image. However, as the robotic pen drawing system uses far fewer stippling points than printing, the result may be reproduced poorly depending on the color spectrum of the input image.
Generating our own color palette instead of using predefined color spaces like CMYK can be an alternative approach. This involves partitioning raster image pixels containing color information into $k$ different groups, which can be achieved using various clustering algorithms (\ie K-Means, Agglomerative, or DBSCAN). Unlike CMYK, the color channels obtained by this method use representative colors from the pixels. 
This may allow for more accurate colors with fewer points. However, a drawback is that the color of the physical drawing tool (pen) may not match the obtained color perfectly. {Consequently, we have adopted the predefined color palette approach since it can maintain a consistent color palette across all images.}

{\bf Robotic Hardware:}
We use two distinct robotic setups, each with unique advantages. The mobile manipulator configuration excels in its capacity to create expansive artworks on a larger scale. However, it requires manual intervention for color changes. On the other hand, the dual-arm setup offers autonomous tool-changing capabilities. Our design choice involves constraining the drawing canvas using the shared reachable range of both arms, which results in each arm drawing sequentially. Alternatively, it is possible to divide the task based on the reachability of each arm, allowing for the creation of larger artworks for both arms drawing simultaneously. In this case, the system may need to duplicate drawing resources, including multiple color palettes and drawing tools, to maximize the drawing concurrency and reachability. This approach is one of our future research directions.

\section{Conclusion} \label{sec:conclusion}

We introduce TSP-Bot, a multi-color robotic pen drawing system that translates digital raster images into continuous paths on a physical surface, replicating the original colors by segmenting the image into predefined color spaces. Our robotic hardware executes this intricate task by following paths comprising thousands of points, suitable for robots with high accuracy and repeatability. Demonstrating its capabilities, we utilize both a dual-arm robotic setup with a color change mechanism and a mobile manipulator setup, showcasing flexibility beyond canvas size. Our system produces colorful and aesthetically appealing TSP pen art.

Future directions involve further path optimization to enhance robotic drawing performance, including investigating the effects of varying parameters like $d_\epsilon$ and $\kappa_\epsilon$ on robot joints and resulting drawings. Additionally, immediate plans include further exploration of the dual-arm drawing setup. This entails distributing the drawing task between the arms based on their reachability, which will also alleviate the limitation of a restricted drawing canvas space. This approach will require an efficient collision-aware motion planning method and a new drawing strategy.

\section*{ACKNOWLEDGMENT}

This work is supported in part by the ITRC/IITP program (IITP-2024-2020-0-01460) and NRF (2022R1A2B5B03001385) in South Korea.




\bibliographystyle{IEEEtran}{}
\bibliography{references}

\end{document}